\documentclass[letterpaper, 10 pt, conference]{ieeeconf}
\IEEEoverridecommandlockouts % for \thanks
\usepackage{graphicx} % for \includegraphics
\usepackage{booktabs}     
\usepackage[colorlinks, citecolor=red]{hyperref}
\usepackage{cite}
\usepackage{float}
\usepackage{comment}
\graphicspath{{./pdf/}} % load figure within subfolder ./pdf/
\DeclareGraphicsExtensions{.pdf}
\usepackage{algorithm}
\usepackage{algorithmic}
\usepackage{hyperref}
\hypersetup{urlcolor=blue}

\renewcommand{\algorithmiccomment}[1]{\bgroup\hfill//~#1\egroup}
\usepackage[export]{adjustbox} % for \adjincludegraphics

\usepackage[cmex10]{amsmath} % assumes amsmath package installed

\usepackage[short]{optidef} % for argmini environment

\usepackage{bm} % for \bm

\usepackage{epsfig} % for postscript graphics files

\usepackage{times} % assumes new font selection scheme installed

\usepackage{color} % for \color
\usepackage{dsfont}
\usepackage{amssymb}  % assumes amsmath package installed
\usepackage{multirow}

\newcommand{\mytilde}{\raise.17ex\hbox{$\scriptstyle\mathtt{‌​\sim}$}}

\title{\LARGE \bf
	Pseudo-label Guided Cross-video Pixel Contrast for Robotic Surgical Scene Segmentation with Limited Annotations
}

\author{Yang Yu, Zixu Zhao, Yueming Jin, Guangyong Chen, Qi Dou and Pheng-Ann Heng % <-this % stops a space
	\thanks{This work was supported by Key-Area Research and Development Program of Guangdong Province, China under Grant 2020B010165004,
Hong Kong RGC TRS Project No. T42-409/18-R and Hong Kong Innovation and Technology Fund Project No. GHP/080/20SZ.}
	\thanks{Y. Yu, Z. Zhao, Q. Dou and P. A. Heng are with the Department of Computer Science and Engineering, The Chinese University of Hong Kong, Hong Kong. P. A. Heng is also with 
Guangdong Provincial Key Laboratory of Computer Vision and Virtual Reality Technology, 
Shenzhen Institutes of Advanced Technology, Chinese Academy of Sciences, Shenzhen, China. Yueming Jin is with the Wellcome/EPSRC Centre for Interventional and Surgical Sciences (WEISS) and the Department of Computer Science, University College London. G. Chen is with Zhejiang Lab.}
	\thanks{\emph{Corresponding author: Yueming Jin (yueming.jin@ucl.ac.uk) }}
}

\begin{document}

\maketitle
\thispagestyle{empty}
\pagestyle{empty}

\begin{abstract}
Surgical scene segmentation is fundamentally crucial for prompting cognitive assistance in robotic surgery. However, pixel-wise annotating surgical video in a frame-by-frame manner is expensive and time consuming. To greatly reduce the labeling burden, in this work, we study semi-supervised scene segmentation from robotic surgical video, which is practically essential yet rarely explored before.
We consider a clinically suitable annotation situation under the equidistant sampling.
We then propose \textit{PGV-CL}, a novel pseudo-label guided cross-video contrast learning method to boost scene segmentation. It effectively leverages unlabeled data for a trusty and global model regularization that produces more discriminative feature representation. 
Concretely, for trusty representation learning, we propose to incorporate pseudo labels to instruct the pair selection, obtaining more reliable representation pairs for  pixel contrast. 
Moreover, we expand the representation learning space from previous image-level to cross-video, which can capture the global semantics to benefit the learning process. 
We extensively evaluate our method on a public robotic surgery dataset EndoVis18 and a public cataract dataset CaDIS. Experimental results demonstrate the effectiveness of our method, consistently outperforming the state-of-the-art semi-supervised methods under different labeling ratios, and even surpassing fully supervised training on EndoVis18 with 10.1\% labeling. Our code is available at \href{https://github.com/yangyu-cuhk/PGV-CL.git}{https://github.com/yangyu-cuhk/PGV-CL}.

\textit{Index Terms}---Scene segmentation, pixel-level contrastive learning, semi-supervised learning, robotic surgical video
\end{abstract}

%======================================================
\section{INTRODUCTION}

With the assistance of medical robot, minimally invasive surgery (MIS) has greatly reshaped patient care, bringing safer surgery procedure and shorter recovery time~\cite{palep2009robotic}. 
Semantic scene segmentation from surgical video is an essential prerequisite for robot-assisted system.
By providing pixel-wise context of instrument and anatomy, it can facilitate cognitive assistance,
% of what anatomy a surgeon is interacting, 
serving as a building block for higher-level perception, such as surgical decision making~\cite{maier2017surgical} and skill assessment~\cite{poursartip2018analysis}.
%largely enhancing the perception of surgeons. 
In addition, whole scene mask can help selectively render different parts in the augmented reality environment, bringing new possibilities for robotic surgery education and navigation~\cite{allan20202018}.
Precise identifying the robotic instruments is also a critical technique for tool pose estimation~\cite{allan20183}, robot control~\cite{du2019patch} and surgical task automation~\cite{nagy2019dvrk}.
%Pixel-wise semantic segmentation for surgical video can enhance surgeon's perception of surgical scene. 
%Segmented mask of robotic instruments can be used for pose estimation, tool tracking and control. 
%The semantic segmentation can also provide mask for augmented reality to reconstruct the surface of tissue. 

Recently, convolutional neural networks have achieved remarkable successes in surgical scene segmentation~\cite{allan20202018,ren2020task,grammatikopoulou2021cadis,wang2021noisy,jin2022exploring} given large amounts of labeled data.
%are available. 
For example, Ren et al.~\cite{ren2020task} decompose the task into different levels and coordinate multi-task learning. 
However, it is time consuming and laborious for experienced surgeons to perform dense pixel-wise frame-by-frame annotations.
%annotate each pixel of the whole scene for all the frames of videos.
In this work, we study semi-supervised scene segmentation by utilizing scarce labeled data and abundant unlabeled data, which is a highly desired task for practical usage yet still underexplored in the robotic surgery scenario.
Meanwhile, it is highly challenging to precisely segment both tool and anatomy in the complicated surgical scene, given the low inter-class variation between tissues, extremely tiny size of some objects such as thread, and inevitable visual occlusion from blood, tool motion blurriness, and lighting changes.
%To reduce the amount of labeled data required, some semi-supervised methods are proposed to utilize limited amount of labeled data and learn from unlabeled data by creating extra supervision signals. 

\begin{figure}[t]
	\centering
	\includegraphics[width=0.45\textwidth]{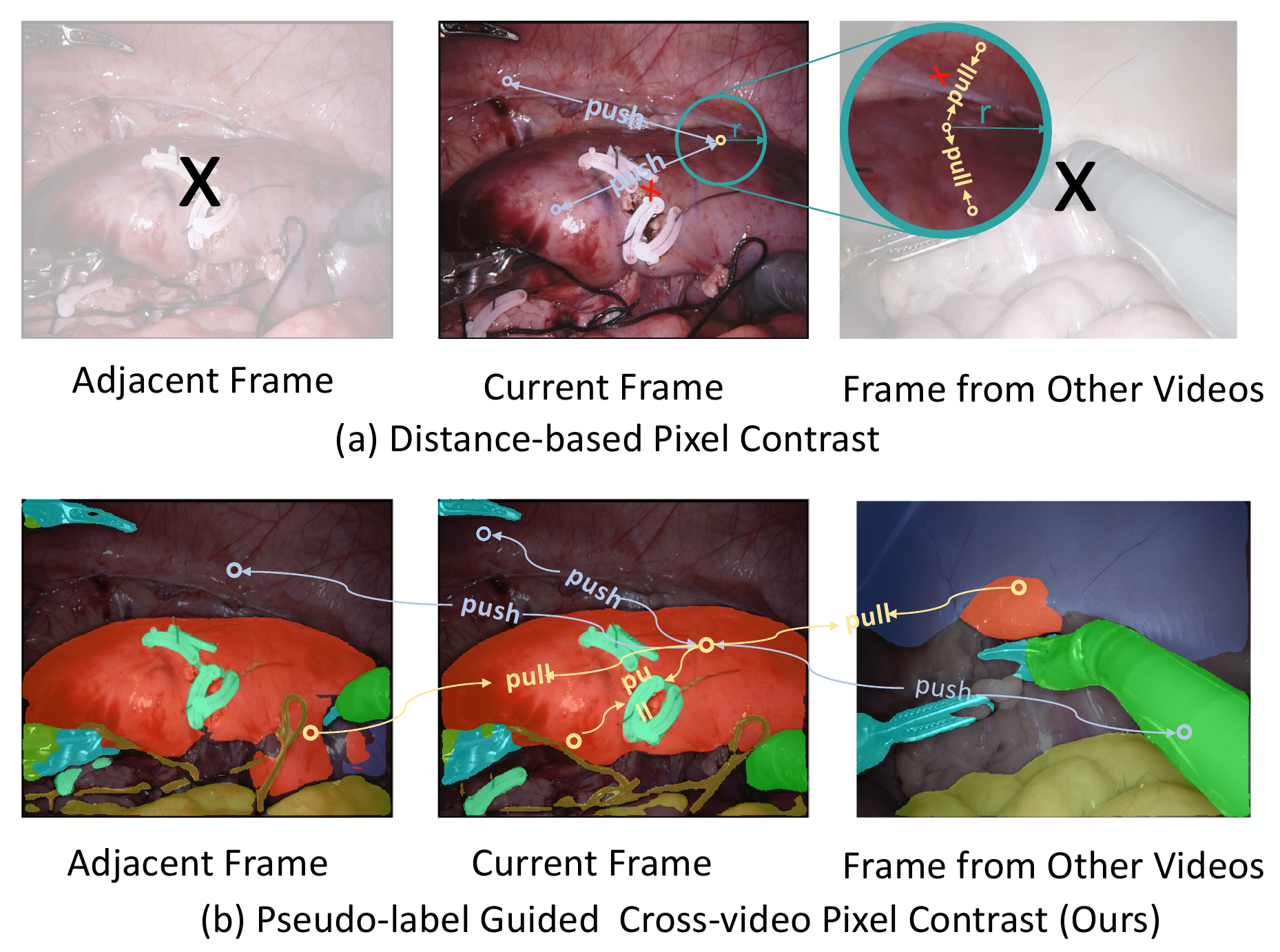}
	\vspace{-3mm}
	\caption{(a) Distance-based pixel contrast can only utilize current frame and pulls (pushes) representations of pixel pair nearby (faraway).  (b)  Our method leverages \textit{pseudo labels} as guidance for trusty contrast, pulling (pushing) representations of pixel pair belonging to the same (different) class. Our pixel pairs are selected \textit{cross-video} to conduct global contrast.}
	\label{intro}
	%\vspace{-6mm}
\end{figure}

Two main streams of approaches have been proposed for semi-supervised semantic segmentation, however, we identify the corresponding limitations of each stream when tackling the surgical scene segmentation with few annotations.
The first stream is generating pseudo labels for unlabeled data by employing the model trained on labeled data, e.g., using segmentation model predictions~\cite{lee2013pseudo, iscen2019label}, or using motion flow to propagate pseudo labels~\cite{jin2019incorporating,zhao2020learning}.
%to provide extra supervision signals.
%Some recent semi-supervised works on robotic surgery use the motion flow to propagate the pseudo labels of instruments for instrument segmentation~\cite{zhao2020learning}, while is unsuitable for the whole scene segmentation.
Recent studies are dedicated to improving the quality, e.g., utilizing model confidence to filter out poor labels~\cite{zou2018unsupervised,hung2018adversarial, rizve2021defense}, or developing several models and leveraging the inter-model disagreement to locate label errors~\cite{feng2020dmt}.
Even though undergoing the error filtering, all these methods explicitly regard pseudo labels as ground truth to calculate loss for model penalization at the end.
It is sub-optimal for analyzing highly complex robotic surgical scene, as the inevitable noise in pseudo labels degrades the model training when using this explicit supervision (see Sec.~\ref{EXPERIMENTS-com}).
How to more effectively leverage the pseudo label for surgical scene segmentation is still an open question to be solved.

\begin{figure*}[t]
	\centering
	\includegraphics[width=0.97\textwidth]{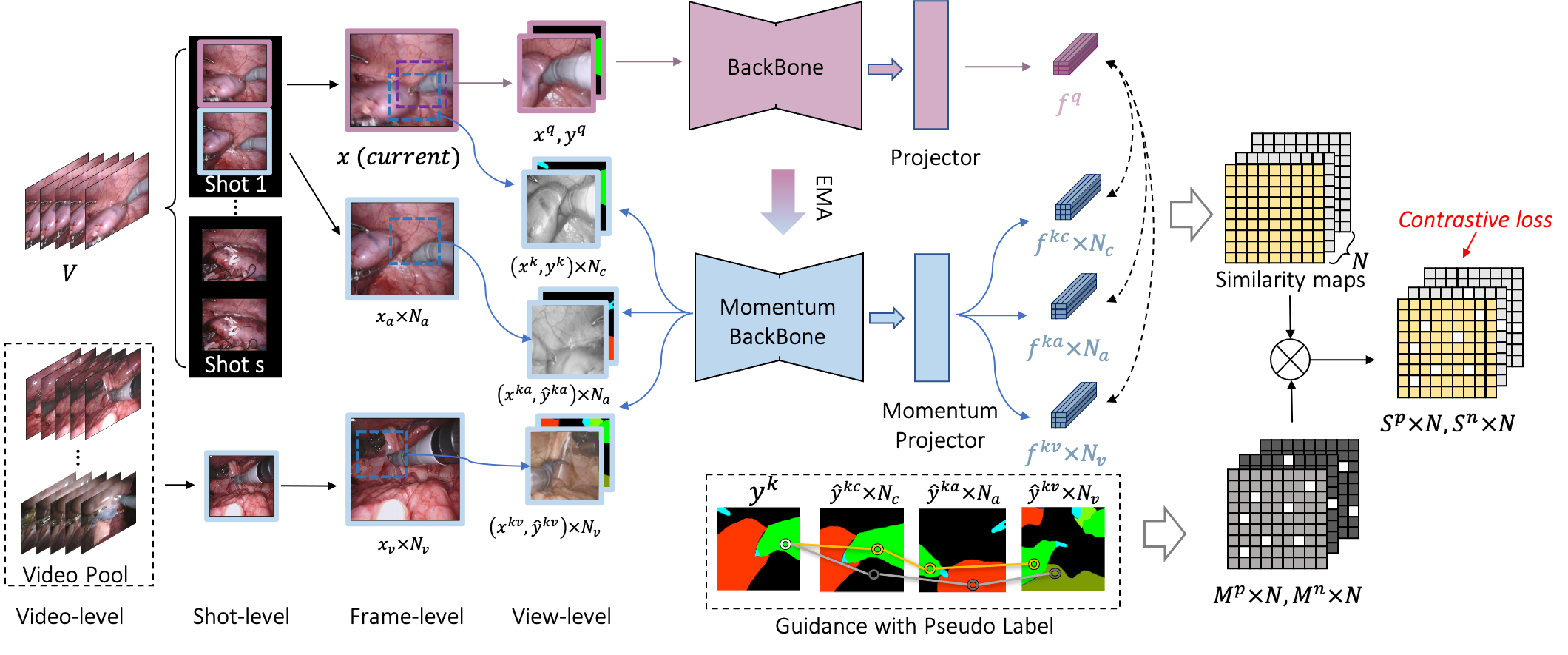}
	\vspace{-4mm}
	\caption{Overview of proposed PGV-CL. It goes through the four-level hierarchy to form the cross-video query $x^q$ and key $\{x^k, \!x^{ka},\! x^{kv}\}$ image samples as the input of model. Dense query and key feature maps are respectively generated by two symmetric branches of the model, for calculating the similarity between per pixel, obtaining one similarity map for each query-key image pair. We then exploit corresponding pseudo labels as the guidance to pair query and key pixels to positive $\mathcal{M}^p$ and negative $\mathcal{M}^n$ ones. Based on these, similarities can be partitioned to $S^p$ and $S^n$ for the pixel-wise contrastive learning.}
	\label{fig:method}
\end{figure*}

Recent advances are dedicated to the other direction, which aims to extract knowledge from unlabeled data through self-supervised regularization.
Consistency regularization methods~\cite{french2019semi, tarvainen2017mean,yu2019uncertainty} are first proposed to regularize the model to produce consistent predictions when inputting an unlabeled data with different perturbations. 
Contrastive learning emerges to regularize the model in a higher-level feature space~\cite{chen2020simple,chen2020improved}. %for model representation learning.
The core idea is to attract similar (positive) and repulse dissimilar (negative) pairs of features extracted from unlabeled data, to provide a good model initialization.
With advanced representation capability, after using limited labeled data for fine-tuning, the model can achieve precise scene segmentation~\cite{chaitanya2020contrastive,lai2021semi}.
One main investigation direction is how to accurately construct the positive and negative pairs, given its key role in contrastive learning ~\cite{chen2020simple,chen2020improved,xie2021propagate,chaitanya2020contrastive}.
%Most state-of-the-art 
Some methods construct pairs based on the pixel location~\cite{chaitanya2020contrastive,xie2021propagate,lai2021semi}. For example, Xie et al.~\cite{xie2021propagate} utilize the spatial distance with the assumption that neighborhood pixels generally belong to the same class and can form the positives.
However, extensive exceptions will appear in the robotic surgical scene. See Fig.~\ref{intro}(a), different tissues and instruments with irregular shape generally present at the same time, bringing massive class boundaries. Pixels near these boundaries are hardly formed to be the precise pairs.
Moreover, most existing methods construct pairs purely from the current single image, ignoring the valuable sequential information in robotic surgical video.

%Its effectiveness can be credited to the smoothness assumption or the cluster assumption that sample points close to each other in the input space should belong to the same class. 
In this work, we propose a novel \textbf{p}seudo-label \textbf{g}uided cross \textbf{v}ideo \textbf{c}ontrast \textbf{l}earning (PGV-CL), to tackle semi-supervised scene segmentation of robotic surgical video.
Our contrastive paradigm can boost the segmentation performance via a \emph{trusty} and \emph{global} regularization. It enhances the model representation learning from two perspectives, i.e., the accuracy and adequacy of the contrast representation space.
Specifically, %our framework wisely integrates the respect advantages of two stream approaches.
we first devise an equidistant sampling strategy for semi-supervised scene segmentation, i.e., only performing the labeling at interval, to maximize the quality of pseudo labels.
Unlike previous methods treating pseudo labels as ground truth for explicit model regularization by loss, we propose to use the pseudo labels as the guidance for pair construction in contrastive learning, sharing the same spirit with very recent works that devise a label-based contrastive loss ~\cite{zhao2020contrastive,zhou2021c3,wang2021exploring}.
Our implicit way to use the pseudo label bypasses accumulating mistakes when treating it as ground truth, and can greatly increase the preciseness in contrastive training.
Moreover, it enables to borrow more knowledge in robotic surgical videos in contrastive training.
Here, we consider two properties: i) robotic surgical video presents the inherent sequential nature of robotic surgical video, where adjacent frames share the similarity semantic information;
ii) different surgical procedures generally present various appearances due to different lighting conditions, patient cohorts, or surgeon operative skills.
In combination with the pseudo label guidance, we propose a cross-video pixel contrast, introducing the more aggressive positive pairs from the adjacent frames, and natural negative pairs from different videos, to enlarge the contrast space.

With the increase in both accuracy and adequacy for contrast, our method can shape a better structure in the global representation space, and providing better model initialization.
To this end, only using extreme limited annotations, our method can achieve accurate segmentation on surgical scene.
Our main contributions are summarized as follows:

\begin{itemize}
	\item We take the first step to integrate the pseudo label into contrastive learning for semi-supervised scene segmentation of robotic surgical video, contributing to the accurate and trusty pixel contrast with guidance.
	\item We develop a novel four-level hierarchical pair construction for cross-video contrast, which leverages the inherent properties of robotic surgical videos, to consider the global semantics of the whole dataset in feature representation learning.
	\item We extensively validate our method on a public robotic surgery dataset EndoVis18~\cite{allan20202018} and a public cataract dataset CaDIS~\cite{grammatikopoulou2021cadis}. Our method consistently outperforms state-of-the-art semi-supervised methods by a large margin, even exceeding the fully supervised training with 10.1\% labels on EndoVis18.
	 
\end{itemize}

%======================================================	
\section{METHODS}
\label{METHODS}
Fig.~\ref{fig:method} presents the overview of proposed PGV-CL for semi-supervised scene segmentation from robotic surgical video.
We first describe the problem setting with devised annotation sampling strategy, which can also provides relatively accurate pseudo labels. 
We then introduce our pseudo-label guided mechanism and cross-video pair construction to better shape the pixel representation space, from two aspects of the accuracy and adequacy of contrast.

%The proposed XXX learns from frames labeled with intervals and generates pseudo labels for the remaining unlabeled frames. Then pseudo labels serve as guidance to construct pos-neg pairs in pixel-level contrastive learning. With pseudo label as guidance, pixel-level contrastive learning is conducted in video-wise. The following subsections describe the problem setup, pseudo-label guided pixel-level contrastive learning, and video-wise positive/negative pairs construction.

\subsection{Equidistant Sampling for Semi-supervised Segmentation}

%We study a semi-supervised scene segmentation for robotic surgical video, where only partial frames have the dense labeling.
Adjacent frames from the surgical video generally share the similar appearance, making model easily parse the scene after seeing one of them.
To maximize the quality of pseudo label under the same annotation efforts, we propose an equidistant sampling strategy, where the annotation is performed sparsely on each training video sequence with an interval.
Given a video having $T$ frames as $ V \!= \!\{x_{0}, x_{1}, ..., x_{T-1}\} $, we assume that only $ \{x_{0}, x_{1\times h},...,x_{t\times h}\} $ are labeled with interval $h$, where $t\times h$ is the largest integer smaller than $T-1$. For example, annotating the surgical video with the interval 9 accounts for 10\% labeled data. With equidistant sampling strategy, our model generates relatively accurate pseudo label for unlabeled frames, as shown in Fig.~\ref{fig:pseudo}. And the pseudo label is accurate enough, thus it is directly used to guide our cross-video pixel contrast.
%thus accounting for $h^{-1}$ labeled data. Fig.~\ref{intro}(a) shows the condition of the interval 9.
%To understand this equidistant sampling strategy, we assume the high-level embedding vector of each video frame to be generated from a Brownian motion given the fact that nearby frames always share similar content. Let $e_t\in\mathbb{R}^d$ denote the embedding vector of the frame $x_t$, then $e_t = e_{t-1}+m_t$ with $m_t\in\mathbb{R}^d$ as the multivariate normal distribution. 
%Performance of our equidistant sampling can be evaluated and theoretically bounded~\cite{janssen2009equidistant}. %, see~\cite{janssen2009equidistant} for details.
This setting is also favorable to clinical practice, as it is easier for surgeons to perform low hertz labeling of surgical video.

With the interval $h$, the whole dataset consists of labeled subset $ \mathcal{D}_{L}=\{(x_{t}, y_{t})\}_{t=hn}$ and unlabeled subset $\mathcal{D}_{U}=\{x_{t}\}_{t \neq hn} $.% with $M = hN$ frames.
We first train a segmentation model $\mathcal{F}^s:\{x_{t}\}\to \{y_{t}\}$ using the labeled subset, and use $f^s$ to generate the pseudo labels, obtaining $\hat{\mathcal{D}}_{U}=\{x_{t}, \hat{y}_{t} \}_{t \neq hn}$.
$\hat{\mathcal{D}}_{U}$ is then utilized for the pseudo label guided contrastive learning.

%In semi-supervised scene segmentation, only a little part of frames are labeled while the remaining frames stay unlabeled. As a kind of sequential data, adjacent frames of a video naturally share similar content. Intuitively, model trained on a frame can give better prediction for frames near it than frames far from it, as shown in Fig. 1 (b). 
%To improve the quality of pseudo labels, we propose to label video frames with intervals. Given a video including T frames as V = \{$F_{0}$, $F_{1}$, ..., $F_{T-1}$\}, we assume that V is labeled with intervals. For example, only $F_{0}$, $F_{10}$, $F_{20}$, ... are labeled with interval 9, resulting in 10\% labeled data, as shown in Fig.~\ref{intro}(a).%Fig. 1 (a). 

\subsection{Pseudo-label Guided Contrast for Trusty Regularization}
\label{sec:pse}
Contrastive learning can enhance the discriminative capability of the model by self-training on unlabeled data.
Given an image $x \! \in \! \mathbb{R}^{H\!\times \!W \!\times\! 3}$ as the query sample and a set of images as key samples, image-level contrastive learning train the model $\mathcal{F}^{CL}$ by distinguishing the positives (augmentation version of $x$) from negatives (from training set excluding $x$).
The contrastive loss is based on the similarity principle between the feature embeddings of these data samples: $f \!= \!\mathcal{F}^{CL} (x), f \!\in\! \mathbb{R}^D$, $D$ is the feature channel.

For the dense surgical scene segmentation, we explore a pixel-level contrastive learning by extending data sample from image to image pixel.
%$f^{CL}$ predicts the dense feature map $f \! \in \! \mathbb{R}^{H\!\times \!W \!\times\! D}$, in which each vector corresponding to each image pixel. 
Concretely, we perform different augmentations on the current frame $x$ and generate two views $x^q$ as query and $x^k$ as key. 
With two symmetric branches of our framework (cf. Fig.~\ref{fig:method}), $x^q$ is fed into a regular encoder branch to predict the query feature map: $f^q \!= \!\mathcal{F}^{CL}_q (x^q)$, and likewise, $x^k$ is fed into a momentum encoder branch to predict the key feature map: $f^k\! = \!\mathcal{F}^{CL}_k (x^k)$. %Both feature maps $f^q,f^k \! \in \! \mathbb{R}^{H\!\times \!W \!\times\! D}$, in which each vector corresponding to each image pixel. 
The encoder branches are composed of a backbone model and a projection head. We employ our pre-trained segmentation model $\mathcal{F}^s$ without classifier as the backbone, to well-initialize most layers. 
Directly applying the contrastive objective onto the features would regularize the representation learning too heavily, we thus construct a projection head to map the features to a lower-dimension space. %follow previous work~\cite{xie2021propagate} to
With the feature maps $f^q,f^k \! \in \! \mathbb{R}^{H\!\times \!W \!\times\! D}$, we can extract the pixel-level feature embeddings, denoting the query embedding as $f^q_i$ and the key embedding as $f^k_j$ for the $i$ and $j$ pixel, respectively.
We then compute the cosine distance on the obtained embeddings:
\begin{equation}
\vspace{-1mm}
\label{eq:1}
S_{ij} = \frac{\langle f^q_i, f^k_j \rangle}{\|f^q_i\| \|f^k_j\|}.
\vspace{-1mm}
\end{equation}
A 2D similarity map $S$ then can be formed by calculation on all pixels, to measure the similarity of the query $x^q$ and key $x^k$ in pixel-level.

\begin{figure}[t]
	\centering
	\includegraphics[width=0.4\textwidth]{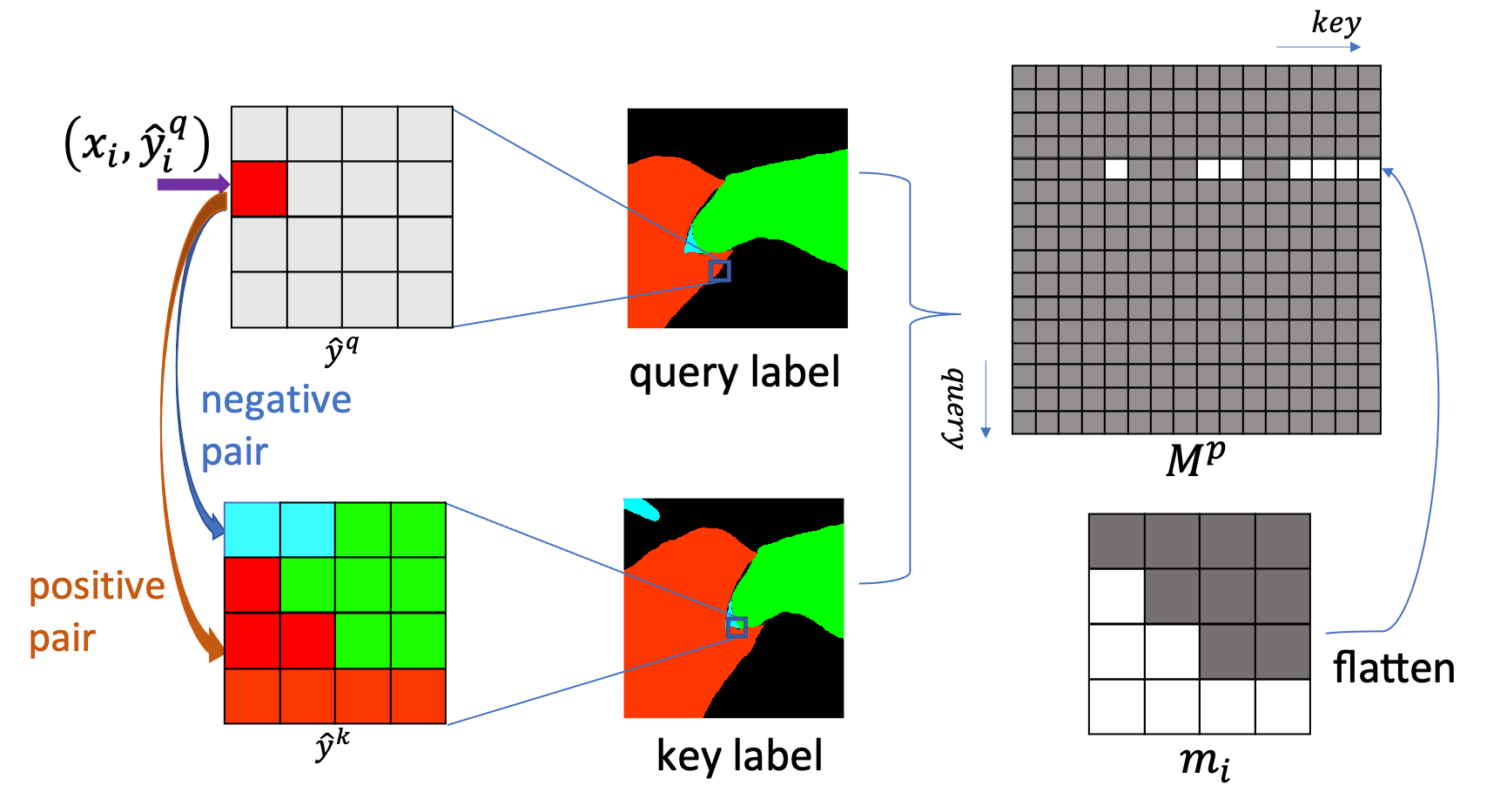}
	\vspace{-3mm}
	\caption{A query pixel $i$ forms positive pairs with pixels from key view having the same labels. Aligned with label $\hat{y}^k$, we get a positive map $m_i$ of pixel $i$ with label $\hat{y}_i^q$. Flatten and stack all maps of pixels in $\hat{y}^q$,
	we get a mask with the same size as similarity map.}
	\label{fig:label_guided}
	%\vspace{-6mm}
\end{figure}

Most existing works relied on the spatial distance between pixels to construct positive and negative pairs~\cite{xie2021propagate}, which easily suffers from the erroneous outcomes for complex surgical scene.
Instead, we pursue a more trustworthy way by leveraging our relatively precise pseudo label to navigate the process of forming pairs.
It is based on the underlying assumption that embeddings of pixels belonging to the same class should be closer than those from different classes. 
Concretely, we perform the same augmentation on pseudo labels, and then couple $\{f^q_i,\hat{y}_i^q\}$ for pixel $i$ in query image and $\{f^k_j,\hat{y}_j^k\}$ for pixel $j$ in key image.
For each key image, we define two 2D label selection masks $\mathcal{M}^p$ and $\mathcal{M}^n$ respectively for positives and negatives.
Both of them are with a binary variable $\mathcal{M}^p, \mathcal{M}^n \subseteq\{0,1\}^{HW\times HW}$, and value at $ij$ is determined by the pseudo label:
\begin{equation}
\vspace{-1mm}
\label{eq:mask}
\mathcal{M}^p_{ij} = \mathds{1}[{\hat{y}_i^q = \hat{y}_j^k}],~~ \mathcal{M}^n_{ij} = 1 - \mathcal{M}^p_{ij}.
\vspace{-1mm}
\end{equation}
For the query pixel $q_i$, a key frame can provide both positive and negative pixel embeddings. If pseudo semantic labels $\hat{y}_i^q$ and $\hat{y}_j^k$ belong to the same class, Eq.~\ref{eq:mask} gives $\mathcal{M}^p_{ij}\! = \!1$ to categorize $k_j$ as the positive embeddings, otherwise, they are regarded as negative embeddings (cf. Fig.~\ref{fig:label_guided}). % with $\mathcal{M}^n_{ij} = 1$.
With the guidance of label selection masks, we divide the similarity for query pixel $q_i$ into the positive and negative portions through the element-wise production between the masks and similarity map:
\begin{equation}
\label{eq:contrast}
\small
S^{p}_{i} =  \frac{1}{| \mathcal{P}_i|}\sum_{j \in \mathcal{P}_i} S_{ij} \! \odot \! \mathcal{M}^p_{ij}, ~~
S^{n}_{i} =  \frac{1}{| \mathcal{N}_i|}\sum_{j \in \mathcal{N}_i} S_{ij} \!\odot \!\mathcal{M}^n_{ij},
\end{equation}
where $\mathcal{P}_i$ and $\mathcal{N}_i$ are pixel embedding collections of positive ($\mathcal{M}^p_{ij} \!=\! 1$) and negative ($\mathcal{M}^n_{ij} \!= \!1$) samples from the key image. Our pseudo label guided contrastive objective is then designed to maximize the masked positive similarities while minimize the masked negative ones (shown by the overall variant in Eq.~\ref{eq:loss}).

\subsection{Cross-video Contrast for Global Regularization}
Many literatures demonstrate that contrastive learning can benefit from more positive and negative samples by creating more views~\cite{chen2020simple,he2020momentum}, which recently have been proved by an expansion and separation theory~\cite{wei2020theoretical}.
Apart from different views of a single frame created by the artificial augmentation (Sec.~\ref{sec:pse}),
robotic surgical videos can inherently provide natural and more aggressive augmentations to construct more positives and negatives.
However, utilizing temporal information for pair construction in pixel-level is difficult, as correspondence between pixels of different frames is unknown due to motion. %In the robotic surgery, different classes even show high variance in the motion degree.
Thanks to our pseudo-label guided strategy, where such correspondence can be more easily located, we can extend our contrastive learning from image-wise to video-wise by selecting pairs from different video frames. 
We present a hierarchically four-level selection strategy for our cross-video contrast learning, which can regularize model by considering global semantic knowledge of the whole dataset.

As shown in Fig.~\ref{fig:method}, conventional image-level set and view-level set contain the original image and its different perturbations (Sec.~\ref{sec:pse}).
We first grow the pair hierarchy with the shot-level, aiming to increase the positives by borrowing the adjacent frames. Note that such adjacent frames are chosen from the same shot, instead of being simply selected from the same video.
As in robotic surgery, visual content generally keeps consistent within a complete action, while shows a large variation when action transaction or some unexpected events happen. % (blooding, lens cleansing, adverse alert) 
A surgical video therefore is presented as a set of several shots, where frames within each show high similarity. With the proceeding of video flow, intensity histogram changes slightly within the same shot and changes severely when the video moves to next shot. 
In this regard, for the query image $x$ from video $V$, we first exploit intensity histogram to partition $V$ to several shots by detecting the severe change with a threshold. We then randomly sample $N_{a}$ frames from the same shot as $x$, and couple them with pseudo labels $(x^{ka},\hat{y}^{ka})$ followed by the same augmentation to generate the key views.
It is then passed through momentum encoder to generate the key embeddings. %mainly contributes to the positive 
%Memory bank of negatives in~\cite{he2020momentum} have been widely used to increase the contrastive learning with the benefit from a large number of negative samples.
We then expand the pair hierarchy to video-level, to leverage frames from other videos to increase negatives.
Specifically, we randomly sample frames from other videos in video pool, resulting in $N_{v}$ frame and pseudo label pairs $(x^{kv},\hat{y}^{kv})$.
They go through same procedures, i.e., augmentation, feature extraction from momentum encoder, to mainly enrich negative key embeddings.

To this end, for the query pixel $i$, we also use Eq.~\ref{eq:mask} on more key pixels (with $\hat{y}^{ka}_{j}$ and $\hat{y}^{kv}_{j}$) to increase the numbers of pseudo label selection masks to $N \! =\! N_{a} \!+\!N_{v}\!+\! N_{c}$, where $N_{c}$ is the number of different views generated from the current frame. %followed the same procedure of Eq.~\ref{eq:mask}. 
%The pixel embedding collection of positive samples $\mathcal{P}_i$ in Eq.~\ref{eq:contrast} can be enhanced by $\mathcal{P}_i \! =\! \mathcal{P}_i^{a} \! \cup \!\mathcal{P}_i^{v} \!\cup\! \mathcal{P}_i^{c}$, where three subsets respectively consist of the positive pixel embeddings provided by adjacent frames $f^{ka}_{ij}$, frames from other videos $f^{kv}_{ij}$, and the current frame with different augmentation $f^{k}_{ij}$. $\mathcal{N}_i$ can also be augmented in the same way. 
The masked positive and negative similarities in Eq.~\ref{eq:contrast} can be augmented with updated $\mathcal{P}_i$ and $\mathcal{N}_i$:
\begin{equation}
\label{eq:4}
\small
\vspace{-1mm}
S^{p}_{i} =  \frac{1}{| \mathcal{P}_i|}\sum_{j \in \mathcal{P}_i} S_{ij} \! \odot \! \mathcal{M}^p_{ij};
%S_i^p = \frac{1}{| \mathcal{P}_i|}\sum_{j \in \mathcal{P}_i} S^{p}_{ij},
\vspace{-3mm}
\end{equation}
\begin{equation}
\label{eq:5}
\small
%S_i^n = \frac{1}{| \mathcal{N}_i^{a}|}\sum_{j \in \mathcal{N}_i^{a}} S^{n}_{ij} + \frac{1}{| \mathcal{N}_i^{v}|}\sum_{j \in \mathcal{N}_i^{v}} S^{n}_{ij} + \frac{1}{| \mathcal{N}_i^{c}|}\sum_{j \in \mathcal{N}_i^{c}} S^{n}_{ij}.
\vspace{-1mm}
S_i^n = \sum_{id=1}^{N} \frac{1}{| \mathcal{N}_i^{id}|}\sum_{j \in \mathcal{N}_i^{id}} S_{ij} \! \odot \! \mathcal{M}^n_{ij}, ~~
\mathcal{N}_i^{id} \! \subset \! \mathcal{N}_i.
\vspace{-0.5mm}
\end{equation}
Here, the updated $\mathcal{P}_i$ contains the three subsets, which respectively consist of the positive pixel embeddings provided by adjacent frames $f^{ka}_{j}$, frames from other videos $f^{kv}_{j}$, and the current frame with different augmentation $f^{kc}_{j}$. $\mathcal{N}_i$ is augmented in the same way. For different query pixels, the number of positive/negative pairs can be largely different. Thus we use average value of similarities instead of summation value to make value of $\mathcal{S}_{i}^{P}$ and $\mathcal{S}_{i}^{N}$ stable . Note that we only average negative similarities in frame-level, with $\mathcal{N}_i^{id}$ denoting the set of key pixels from one frame. Compared with the positive one that is further averaged all pixels cross frames, this strategy can enlarge the value of negative proportion to benefit the contrastive learning.

%Our overall contrastive objective is defined to maximize positive similarities while minimize negative ones:
We devise a new contrastive objective to maximize positive similarities while minimize negative ones:
\begin{equation}
\label{eq:loss}
%\vspace{-2mm}
\mathcal{L} =  -\frac{1}{k} \sum_{i=1}^{k} \log \frac{\exp (S_i^p)}{\exp (S_i^p)+ \exp (S_i^n)}.
%\vspace{-0.2mm}
\end{equation}
Note that in the learning process, only parameters of regular encoder $\theta$ are updated online via back propagation, while the momentum one  $\theta^m$ to calculate the key are updated slowly by exponential moving average (EMA)~\cite{hunter1986exponentially}: $\theta^{m} \leftarrow m \theta^{m} + (1-m) \theta$, where $m$ is a momentum coefficient. Therefore, they provide relatively stable learning targets for the regular encoder to query.
By making full use of similarities among the pixels, our contrastive learning can better shape the pixel embedding space.
The well-structured space enables the model to only use limited annotations to achieve accurate segmentation.% on surgical scene.

%======================================================

\begin{table*}[]
\centering
\caption{Results of different methods on EndoVis18 dataset for scene segmentation.}
\vspace{-2mm}
\begin{tabular}{c|c|c|c|c|c|c|c}
\hline
\multirow{2}{*}{Methods} & \multicolumn{2}{c|}{Frames used}  & Overall                   & Sequence1                 & Sequence2                 & Sequence3                 & Sequence4                 \\ \cline{2-8} 
                         & Labeled Ratio & Unlabeled Ratio & mIoU(\%)                  & mIoU(\%)                  & mIoU(\%)                  & mIoU(\%)                  & mIoU(\%)                  \\ \hline
DeepLabV3 (baseline)~\cite{chen2017rethinking}    & 100\%           & 0                & 58.48$ \pm $0.43          & 63.86$ \pm $0.59          & 54.98$ \pm $0.98          & 80.81$ \pm $0.23          & 34.25$ \pm $1.26          \\ \hline
DeepLabV3 (baseline)~\cite{chen2017rethinking}    & 5.4\%            & 94.6\%             & 53.72$ \pm $0.60          & 55.78$\pm$0.45            & 51.45$ \pm $0.67          & 75.39$ \pm $0.46          & 32.28$ \pm $0.85          \\
DMT~\cite{feng2020dmt}                      & 5.4\%            & 94.6\%             & 53.70$ \pm $0.16          & 56.23$ \pm $0.20          & 50.78$ \pm $0.16          & 74.42$ \pm $0.27          & 33.37$ \pm $0.51          \\
PixPro ~\cite{xie2021propagate}                   & 5.4\%            & 94.6\%             & 55.47$ \pm $1.08          & 58.66$ \pm $1.45          & 52.26$ \pm $1.23          & 78.42$ \pm $0.76          & 32.50$ \pm $2.31          \\
UA-MT~\cite{yu2019uncertainty}                    & 5.4\%            & 94.6\%             & 56.15$ \pm $0.38          & 56.65$ \pm $1.13          & 52.73$ \pm $0.84          & 78.97$ \pm $0.60          & 36.23$ \pm $0.99          \\
PGV-CL(Ours)             & 5.4\%            & 94.6\%             & \textbf{57.83$ \pm $0.23} & \textbf{59.78$ \pm $0.50} & \textbf{53.37$ \pm $0.60} & \textbf{80.66$ \pm $0.15} & \textbf{37.50$ \pm $1.89}  \\ \hline
DeepLabV3 (baseline)~\cite{chen2017rethinking}    & 10.1\%            & 89.9\%             & 56.76$ \pm $0.30          & 61.07$ \pm $0.38          & 52.69$ \pm $0.19          & 78.99$ \pm $0.15          & 34.27$ \pm $1.17          \\
DMT~\cite{feng2020dmt}                      & 10.1\%            & 89.9\%             & 56.34$ \pm $0.11          & 60.44$ \pm $0.24          & 52.30$ \pm $0.13          & 79.43$ \pm $0.26          & 33.14$ \pm $0.24          \\
PixPro ~\cite{xie2021propagate}                  & 10.1\%            & 89.9\%             & 57.38$ \pm $0.42          & 61.90$ \pm $1.92          & 54.66$ \pm $0.85          & 80.83$ \pm $0.49          & 32.13$ \pm $1.45          \\
UA-MT ~\cite{yu2019uncertainty}                    & 10.1\%            & 89.9\%             & 57.21$ \pm $0.63          & 58.02$ \pm $0.56          & \textbf{55.06$ \pm $0.48} & 79.87$ \pm $0.85          & \textbf{35.86$ \pm $1.13} \\
PGV-CL(Ours)             & 10.1\%            & 89.9\%             & \textbf{58.59$ \pm $0.14} & \textbf{62.70$ \pm $0.58} & 54.53$ \pm $0.87          & \textbf{81.32$ \pm $0.57} & 35.81$ \pm $1.68          \\ \hline
\end{tabular}
\centering
\label{tab:endo18}
\end{table*}

\begin{table*}[]
\centering
\caption{Results of different methods on CaDIS dataset for scene segmentation.}
\vspace{-2mm}
\begin{tabular}{c|c|c|c|c|c|c|c|c}
\hline
\multirow{2}{*}{Methods} & \multicolumn{2}{c|}{Frames used} & \multicolumn{3}{c|}{Validation set}                                               & \multicolumn{3}{c}{Test set}                                                                \\ \cline{2-9} 
                         & Labeled       & Unlabeled      & mIoU(\%)                  & PA(\%)                    & PAC(\%)                   & mIoU(\%)                   & PA(\%)                    & \multicolumn{1}{l}{PAC(\%)}          \\ \hline

DeepLabV3~\cite{chen2017rethinking}               & 100\%            & 0              & 86.12$ \pm $0.17          & 94.28$ \pm $0.05          & 92.38$ \pm $0.17        & 82.47$ \pm $0.07          & 93.29$ \pm $0.04          & \multicolumn{1}{l}{88.72$ \pm $0.13} \\ \hline
DeepLabV3~\cite{chen2017rethinking}               & 1.9\%              & 98.1\%           & 78.99$ \pm $0.15          & 92.05$ \pm $0.08          & 86.13$ \pm $0.21          & 75.50$ \pm $0.17          & 91.88$ \pm $0.07          & 83.11$ \pm $0.07                     \\
DMT~\cite{feng2020dmt}                      & 1.9\%              & 98.1\%           & 80.81$ \pm $0.18          & 92.85$ \pm $0.16          & \textbf{87.97$ \pm $0.10}          &78.67$ \pm $0.27         & 92.75$ \pm $0.22         & \textbf{87.40$ \pm $0.38}                     \\
PixPro ~\cite{xie2021propagate}                  & 1.9\%              & 98.1\%           & 77.60$ \pm $1.10          & 92.92$ \pm $0.13          & 84.00$ \pm $1.13          & 74.38$ \pm $0.04          & 92.44$ \pm $0.15          & 81.49$ \pm $0.12                     \\
UA-MT ~\cite{yu2019uncertainty}                   & 1.9\%              & 98.1\%           & 77.45$ \pm $0.16                   & 92.54$ \pm $0.18                   & 84.31$ \pm $0.40                   & 76.02$ \pm $0.84                   & 92.08$ \pm $0.23                   & 83.46$ \pm $0.83                              \\
PGV-CL(Ours)             & 1.9\%              & 98.1\%           & \textbf{81.57$ \pm $0.48} & \textbf{93.82$ \pm $0.08} & 87.88$ \pm $0.37 & \textbf{78.81$ \pm $0.07} & \textbf{93.13$ \pm $0.06} & 85.73$ \pm $0.00            \\ \hline
DeepLabV3~\cite{chen2017rethinking}              & 2.3\%              & 97.7\%           & 80.76$ \pm $0.09          & 92.70$ \pm $0.04          & 87.73$ \pm $0.06          & 79.15$ \pm $0.08          & 92.54$ \pm $0.04          & 86.29$ \pm $0.11                     \\
DMT~\cite{feng2020dmt}                     & 2.3\%              & 97.7\%           & 81.12$ \pm $0.11          & 93.15$ \pm $0.05          & 87.90$ \pm $0.21          & 79.58$ \pm $0.13          & 92.85$ \pm $0.08          & 87.55$ \pm $0.29                     \\
PixPro ~\cite{xie2021propagate}                  & 2.3\%              & 97.7\%           & 80.72$ \pm $1.14          & 93.26$ \pm $0.03          & 87.25$ \pm $1.35          & 79.14$ \pm $0.39          & 92.87$ \pm $0.10          & 86.05$ \pm $0.64                     \\
UA-MT ~\cite{yu2019uncertainty}                    & 2.3\%              & 97.7\%           & 81.97$ \pm $0.13          & 93.09$ \pm $0.17          & 88.80$ \pm $0.51          & 79.24$ \pm $0.28          & 92.64$ \pm $0.10           & 86.11$ \pm $0.38                     \\
PGV-CL(Ours)             & 2.3\%              & 97.7\%           & \textbf{83.75$ \pm $0.85} & \textbf{94.04$ \pm $0.23} & \textbf{90.09$ \pm $0.73} & \textbf{81.49$ \pm $0.34} & \textbf{93.48$ \pm $0.14} & \textbf{87.85$ \pm $0.37}            \\ \hline
\end{tabular}
\label{tab:cadis}
\end{table*}

\section{EXPERIMENTS}
\label{EXPERIMENTS}
We first introduce two datasets and metrics used to evaluate our method. And then we report implementation details of our method. Further, we compare our methods with other semi-supervised segmentation methods. At last, we ablate two important factors that influence our method.

\subsection{Datasets and Evaluation Metrics}

%We validate our method on two surgical video scene segmentation datasets, i.e. EndoVis18 and CaDIS.

\textbf{EndoVis18}: The EndoVis18 is a public challenge dataset from 2018 MICCAI Robotic Scene Segmentation challenge~\cite{allan20202018}. This dataset consists of 19 sequences among which 15 sequences are for training and 4 sequences for testing. %Each training sequence includes 149 frames while each testing sequence includes 249 frames. 
All sequences are recorded on da Vinci X or Xi system during porcine training procedure. 
Each frame is of a high resolution of 1280$ \times $1024. 
The dataset parses the scene into 12 classes, including different robotic instruments and anatomies.
%According to the requirement of the challenge, we divide pixels of surgical images into 12 classes, including a set of medical instrument classes and a set of anatomical classes. 
%This semantic segmentation task is difficult since different anatomical classes can be similar in appearance.

\textbf{CaDIS}: The CaDIS dataset is a public benchmark focusing on the eyes of patients during the cataract surgery~\cite{grammatikopoulou2021cadis}. 
%generated from the training videos released for CATARACTS challenge~\cite{al2019cataracts} and 
It consists of 25 videos, with 19 videos as training set, 3 videos as validation set and 3 videos as test set. 
All videos are recorded using a 180I camera mounted on an OPMI Lumera T microscope and each frame has a high resolution of 960$ \times $540.
The scene is divided into 8 classes with 4 classes for anatomical tissues, 1 class for all instruments, and 3 classes for all other objects appearing in the scene.

For fair comparison, we follow EndoVis18 challenge to use mean intersection-over-union (mIoU) as metrics to evaluate methods. Overall and results on four test sequences are reported. 
Also following CaDIS benchmark, we employ mIoU, Pixel Accuracy (PA), Pixel Accuracy per Class (PAC) as metrics and report results on both validation and test set.

\subsection{Implementation Details}
We use DeepLabV3~\cite{chen2017rethinking} as our segmentation model and ResNet-50~\cite{he2016deep} pre-trained on ImageNet~\cite{deng2009imagenet} as encoder.
We employ a three-stage training strategy to train our model $\mathcal{F}^s$, i.e., pre-training, contrastive learning, and fine-tuning. 
We first train the segmentation model using the labeled data $ \mathcal{D}_{L}$. The pre-trained model generates pseudo labels for unlabeled frames and serves as model initialization for contrastive learning. 
Next, we perform the proposed contrastive learning paradigm to extract the knowledge by adding unlabeled data $ \mathcal{D}_{L}\! \cup \!\hat{\mathcal{D}}_{U}$ (pseudo labels replaced by ground truths for labeled data).
We take all layers of $\mathcal{F}^s$ except the last two layers (used as classifier) as backbone, and develop a projector consisting of two $1 \!\times\! 1$ convolutional layers to reduce the channels.
Finally, we reuse the labeled data $ \mathcal{D}_{L}$ to fine-tune our segmentation model $\mathcal{F}^s$. We exploited Online Hard Example Mining (OHEM)~\cite{shrivastava2016training} as loss objective in this stage.

\begin{figure*}[t]
	\centering
	\includegraphics[width=0.96\textwidth]{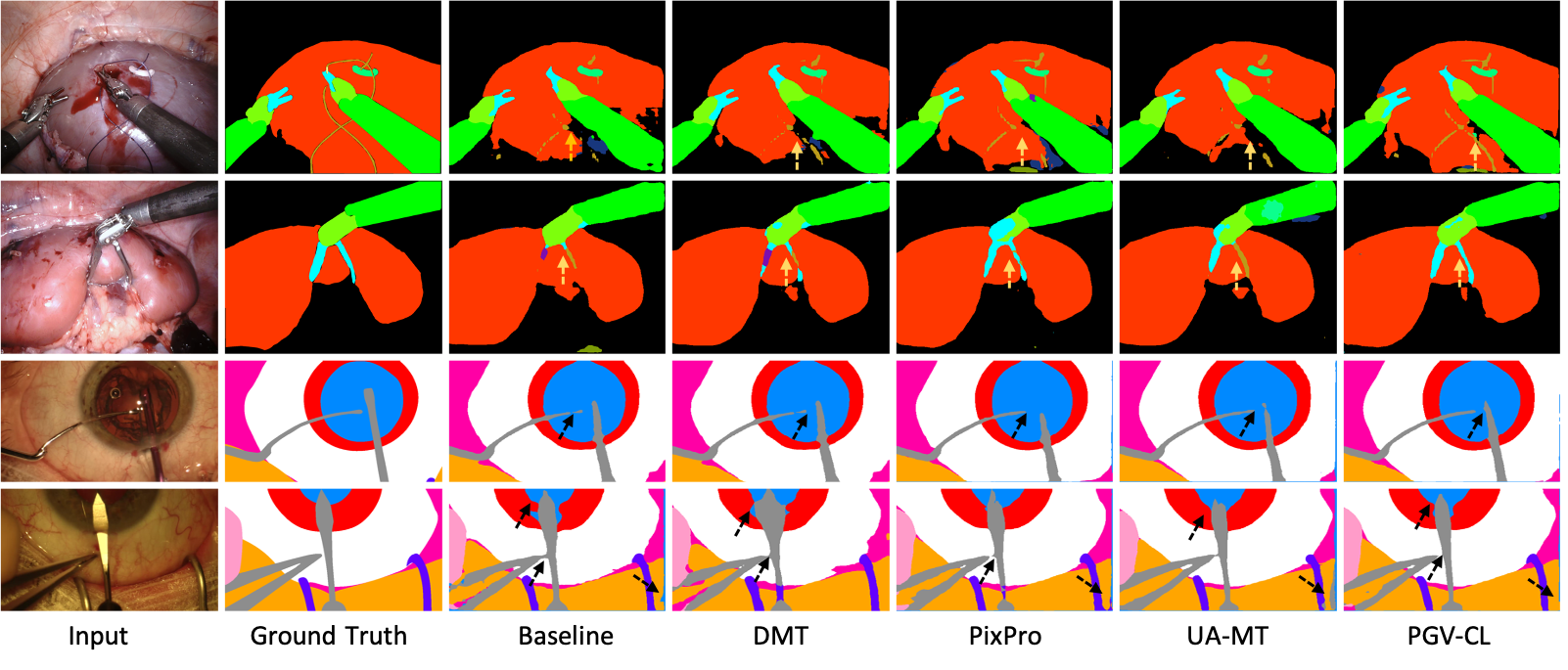}
	\vspace{-2mm}
	\caption{Qualitative comparison on 10.1\% labeled EndoVis18 and 2.3\% labeled CaDIS datasets. More results can be found in supplementary video.}
	\label{fig:visualization}
	\vspace{-4mm}
\end{figure*}

\begin{figure}[t]
	\centering
	\includegraphics[width=0.4\textwidth]{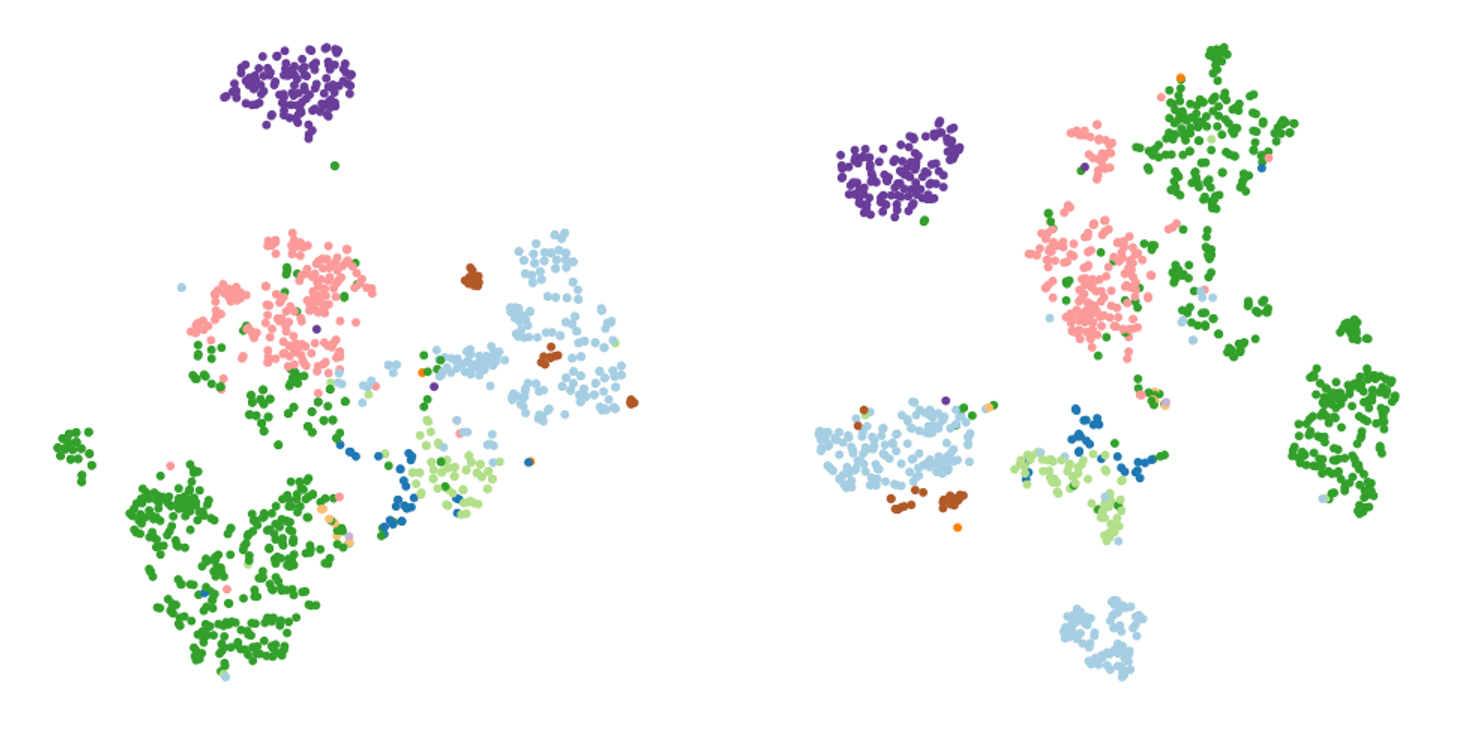}
	\vspace{-5mm}
	\caption{Pixel embeddings from baseline w/o contrast (left) and ours w/ contrast (right) on EndoVis18. Different colors represent different classes.}
	\label{fig:tsne}
	\vspace{-5mm}
\end{figure}

Our framework is implemented in PyTorch with two NVIDIA Titan Xp GPUs. 
All video frames are resized to the resolution of $480 \!\times \! 480$ in the pre-training and fine-tuning stages.
We deploy SGD optimizer~\cite{ruder2016overview} with a poly learning rate scheduler and use a base learning rate of 1e-3 for optimize supervised loss in these two stages. Batch size is 8.
In contrastive learning, we randomly crop a patch (resolution ranging from $144 \!\times\! 144$ to $336\! \times \! 336$) from a frame to create more aggressive augmentation and resize it to $480 \!\times \!480$ to match the resolution used in fine-tuning. 
We use LARS optimizer~\cite{you2017large} with a cosine learning rate scheduler and set a base to 1. %where the learning rate is linearly scaled with the batch size: $lr = lr_{base} \times bs / 256$.  
Batch size is 4 and weight decay is 1e-5. 
Optimizing contrastive loss takes around 7 hours for EndoVis18 and 11 hours for CaDIS with 140 epochs.

\subsection{Comparison with State-of-the-art Methods}
\label{EXPERIMENTS-com}

% \usepackage{multirow}

% \usepackage{multirow}

We implement several state-of-the-art semi-supervised methods for comparison, including \textit{DMT}~\cite{feng2020dmt}: a pseudo-labeling based method, which deploy two differently initialized models to generate pseudo labels for mutual training; \textit{PixPro}~\cite{xie2021propagate}: a contrastive learning method, which form positive and negative pairs based on the spatial distance; \textit{UA-MT}~\cite{yu2019uncertainty}: a consistency regularization method based on the mean teacher, which jointly train the model by segmentation loss and consistency loss.
For fair comparison, all these use the same segmentation model as our method (i.e. DeepLabV3 with ResNet-50). And we implement these methods based on the officially released code, utilize the pre-trained model from ImageNet, and tune some important hyper-parameters on our datasets to get better results. 
Note that for Pixpro, we only replace the contrastive learning stage with other stages remaining the same as our method.
Additionally, we provide a \textit{Baseline} model, that is trained only using the labeled data.
All the experiments are repeated 3 times to avoid the case of coincidence and their average results are reported. 

%We compare our method with baseline and some other semi-supervised segmentation methods, including (i) \textit{Baseline}: our segmentation network trained only using the labeled data. (ii) \textit{PixPro}~\cite{xie2021propagate}: As done in our method, we use PixPro to train DeepLabV3 except its last two convolution layers. To get better results, we use DeepLabV3 pre-trained with labeled frames as initialization of PixPro and adjust distance ratio which is important as reported in their paper. All other training settings are the same as reported in their paper and fine-tuning is the same as our method. (iii) \textit{UA-MT} ~\cite{yu2019uncertainty}: UA-MT is a consistency regularization based method. They train two models by minimizing a segmentation loss and a consistency loss. %To make a fair comparison, we use OHEM as segmentation loss function. 
%We fine-tune the training epoch and all other training hyper-parameters are set as reported in their paper. (iv)  We train our model with hyper-parameters reported in their paper and adjust the training epoch to get better results. All the experiments are repeated 3 times to avoid the case of coincidence and their average results are reported. 

%i.e. 120/2115 and 225/2010 of EndoVis18, 67/3483 and 80/3470 of CaDis. 
Our experiments are conducted on very limited annotation settings, i.e., 5.4\% and 10.1\% frames being labeled with intervals as 20 and 10 on EndoVis18.
Given CaDIS is collected at a higher frequency than EndoVis18 (30 fps v.s. 1 fps), %we validate our method under more challenging settings, i.e., 
only 1.9\% and 2.3\% frames are labeled with intervals of 60 and 50.
The segmentation results on the two datasets are summarized in Table~\ref{tab:endo18} and Table~\ref{tab:cadis}.
We found that by effectively leveraging the unlabeled data, our method can greatly improve the performance compared with the baseline model, i.e., increasing mIoU ranging from 1.83\% to 4.11\% on the four settings.
Among the semi-supervised methods, regarding pseudo label as ground truth to explicitly calculate the loss for model penalization, DMT even degrades results compared with baseline in some extreme annotation settings.
Instead, our method leverages the pseudo label to guide the contrastive learning, largely boosting the segmentation performance.
Meanwhile, compared with Pixpro which relies on the pixel distance for pair construction, the goal of both methods is to pull (push) representations of pixel pair belonging to the same (different) class. Our method achieves superior results, indicating that it can provide more reliable and global model regularization by using pseudo labels to navigate the contrast of pixels across videos. 
%Pixpro and UA-MT achieve the better performances as they extract more reliable information from the unlabeled data.
Additionally, our method consistently outperforms the method that performs the consistency regularization on the model predictions (UA-MT), across all the four settings on two datasets.
The improvement degree is larger as the amount of annotations decrease.
We gain the maximum benefits in the severest condition (1.9\% labeling on CaDIS), with 4.12\% and 2.79\% mIoU increase over UA-MT on validation and test set.
Notably, our method trained with only partly labeled data can even surpass the fully supervised model in some case (10.1\% labeling on EndoVis18).

We show the visual comparison in Fig.~\ref{fig:visualization}. Our method can detect small objects such as thread in row 1 and instrument tips in row 2, and also achieve more complete prediction for large one.
Fig.~\ref{fig:tsne} visualizes the pixel embeddings extracted from baseline and our model with t-SNE~\cite{van2008visualizing}. As pixel-wise features are enormous for all test data, we select five typical frames to cover all classes. We can see that our contrastive learning enables a more separable feature space. Some pixels in the same class although may be divided into two clusters given that they're from different objects and show different appearance, they are more compact for easy distinguishing.

\subsection{Ablation Studies}

\begin{figure}[t]
	\centering
	\includegraphics[width=0.45\textwidth]{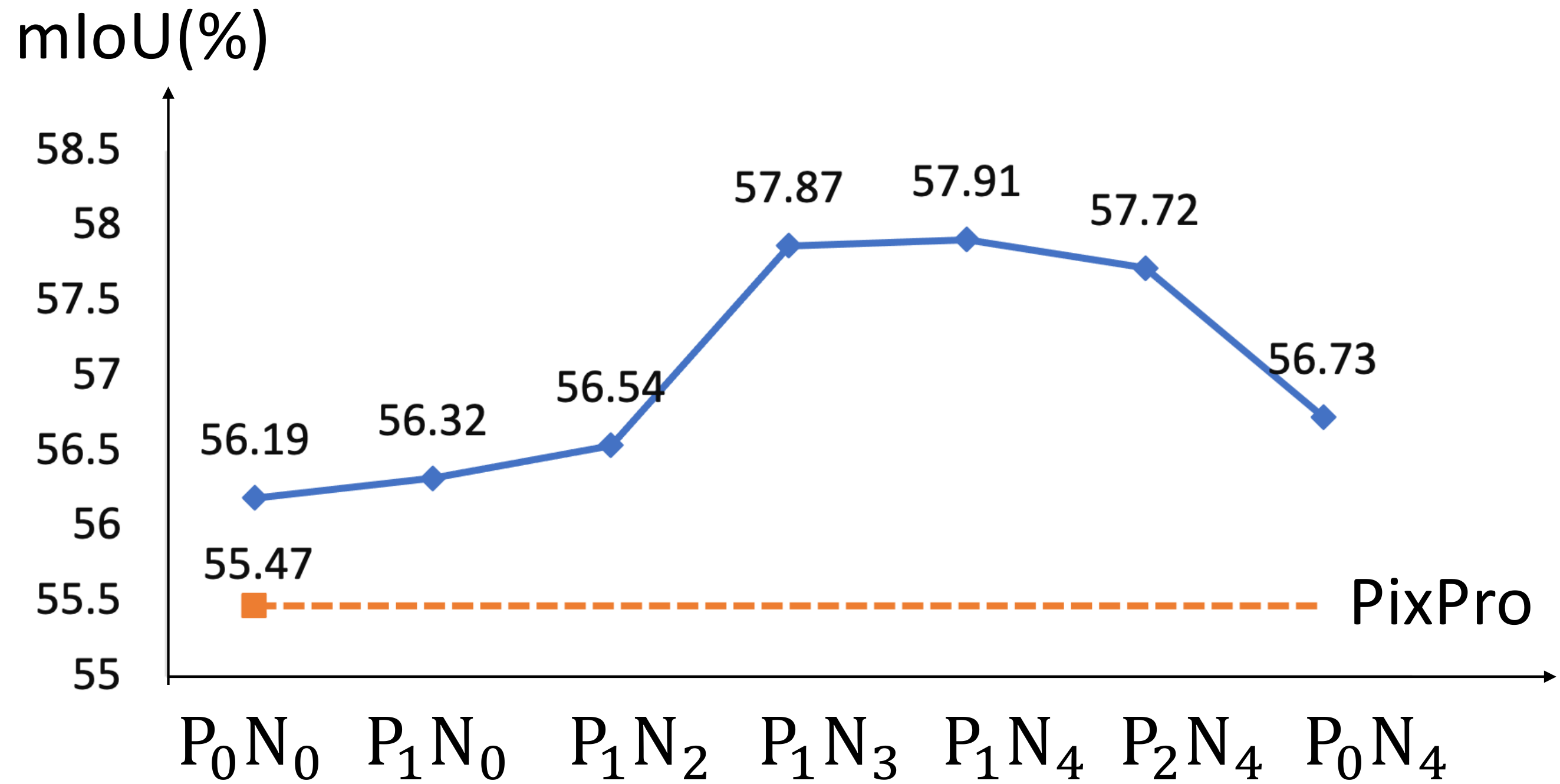}
	%\vspace{-3mm}
	\caption{Ablation study on different amounts of positive and negative pairs.}
	\label{fig:ab1}
\end{figure}

\begin{figure}[t]
	\centering
	\includegraphics[width=0.49\textwidth]{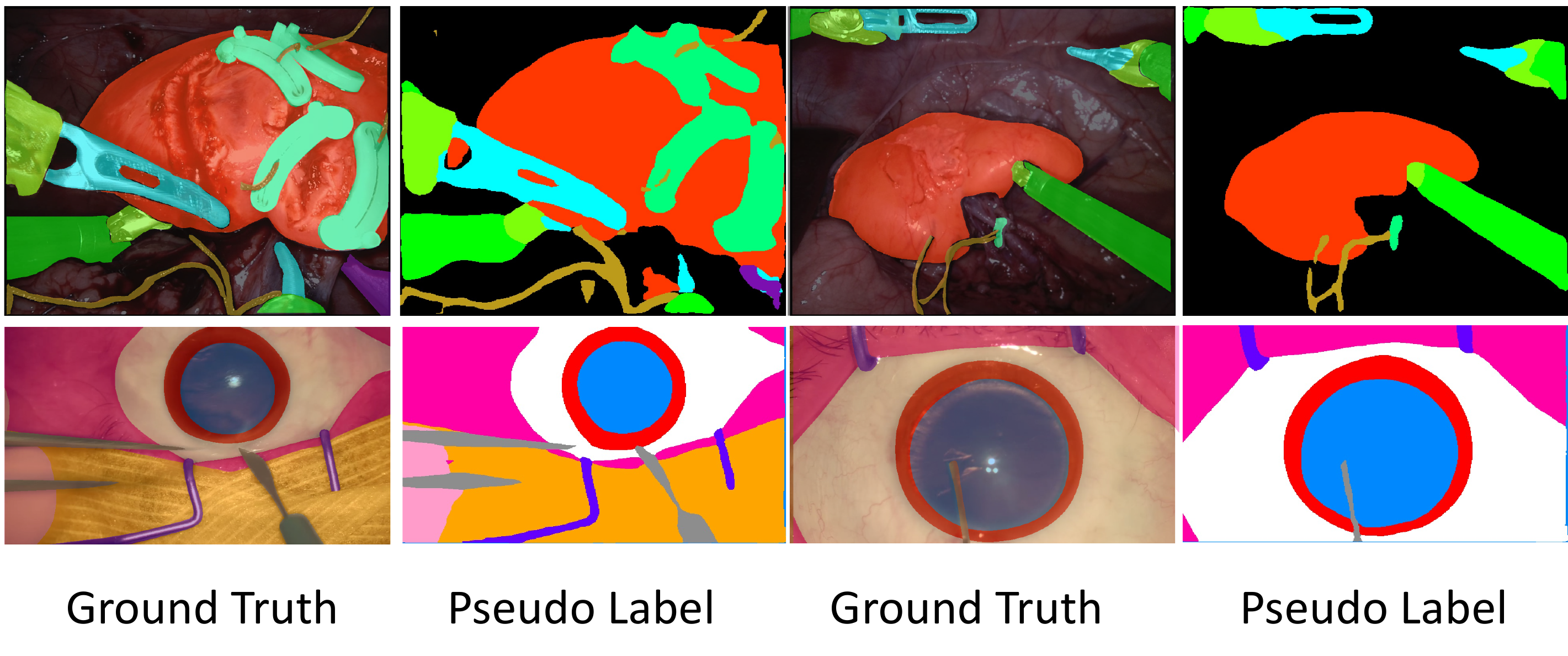}
	\vspace{-5mm}
	\caption{Visualization of generated pseudo labels.}
	\label{fig:pseudo}
	\vspace{-3mm}
\end{figure}

\begin{figure}[t]
	\centering
	\includegraphics[width=0.4\textwidth]{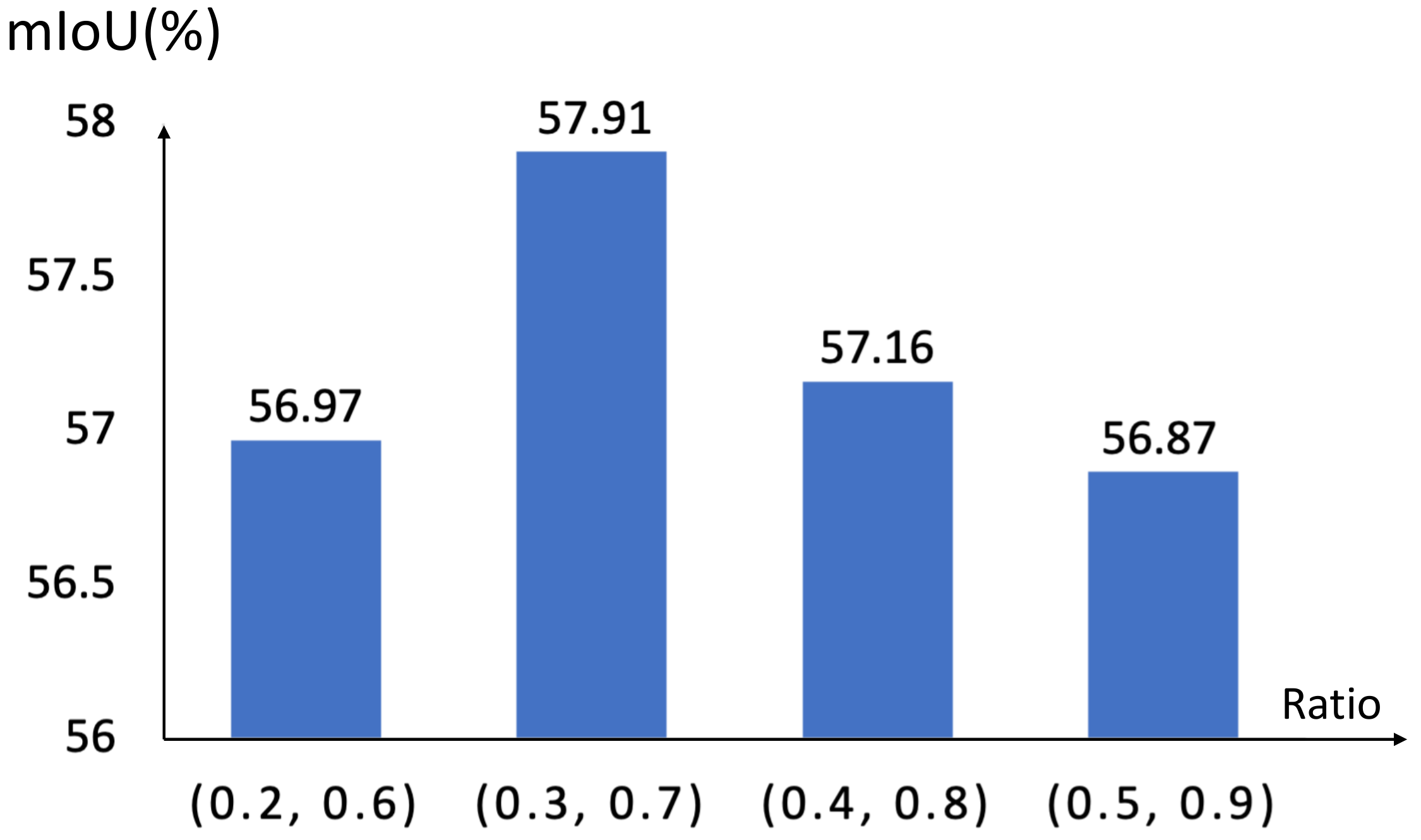}
	\vspace{-3mm}
	\caption{Ablation study on patch crop size in data augmentation.}
	\label{fig:ab2}
	\vspace{-3mm}
\end{figure}

\subsubsection{Positive and negative pair construction}
The pair construction is a key factor in contrastive learning. We analyze its effect on the segmentation results under different qualities, i.e., formed based on spatial distance (Pixpro) and formed guided by pseudo labels (Ours).
We further study its effect with different quantities by varying the value of positive and negative pairs, resulting in six configurations.
With each current frame as the query frame, $P_{i}N_{j}$ means that the numbers of adjacent frames and frames from other videos which we utilize as key frames are $i$ and $j$, respectively.
All experiments are conducted on EndoVis18 dataset with 5.4\% labeled frames.
The results from a single experiment running for each setting are presented in Fig.~\ref{fig:ab1}.

We can see that when the query and key feature maps solely come from a current frame without leveraging the others ($P_{0}N_{0}$), our method increase 0.72\% over the Pixpro, %2.47\% mIoU over the baseline and
demonstrating that our pseudo label guided strategy promotes the precise on pair construction for better contrastive learning.
We further visualize our generated pseudo labels in Fig.~\ref{fig:pseudo} for more intuitive explanation. We can see that they are of high quality and can largely overlap with the ground truths, therefore providing the accurate instruction for pair selection.
Moreover, our pseudo label strategy enables to construct pairs by leveraging other frames, while Pixpro fails to take advantage of it.
Comparing results of $P_{0}N_{4}$, $P_{1}N_{4}$ we see that the adjacent frame mainly providing positive pairs can greatly help achieve the superior performance.
When increasing the value of $j$ to introduce more negative pairs ($P_{1}N_{2}$, $P_{1}N_{3}$, and $P_{1}N_{4}$), segmentation results gradually improve while show a slow trendy when it is larger than 3.
The performance degrades with $P_{2}N_{4}$, given that more positives need more negatives to optimize. 
But including too many negatives is unfeasible, as we do need to consider the practical computation resource for pair construction.
%but too many positives and negatives cause large computation cost.
%The result of $P_{2}N_{4}$ indicates that increasing positives may cause performance degradation. 
%The underlying reason is that more positives need more negatives to optimize, but too many positives and negatives cause large computation cost.
In our work, we therefore choose one adjacent frame and four frames from other videos in pair construction.

%We conduct five configurations to validate the effectiveness of video-wise positive/negative construction. All configurations are evaluated in Endo Vis18 dataset with 5\% labeled frames. In each configuration, we compare a different number of frames used to generate key feature maps for a query feature map. Thus each pixel feature vector forms different amounts of positive/negative pairs. For each current frame, PiNj stands for i adjacent frames and j frames from other videos are used. P0N0 means that no other frames are used and query/key feature map both come from the current frame, which is a frame-wise contrastive learning. Fig.4 shows that increasing the amounts of positive/negative pairs generally results in better performance. From the comparison of P0N4 and P1N4, we can see that positive pairs from adjacent can greatly improve the performance. The comparison of P1N2, P1N3 and P1N4 shows that more negative pairs are needed, but the performance increases slowly with j when j is larger than 3.

\subsubsection{Crop size of patches}
We then analyze the patch crop size, an important augmentation hyperparameter in our contrastive learning.
Specifically, we vary the minimal (\textit{min}) and maximal (\textit{max}) side length in the range of $(0,1]$.
The results of different (\textit{min}, \textit{max}) values on EndoVis18 dataset with 5.4\% labeling are shown in Fig.~\ref{fig:ab2}.
Overall, our method is not very sensitive to this parameter, achieving the stable mIoU around 57\%.
Analyzing in detail, we observe that the higher performance can be achieved when increasing \textit{min} from 0.2 to 0.3.
While when \textit{min} $>0.3$, the results starts decreasing.
The underlying reason is that the crop size directly influences the overlapped area and the amount of formed positive pairs.
Too scarce or redundant positive pairs may lead to the training difficulty in contrastive learning.

\section{CONCLUSIONS AND FUTURE WORK}
\label{CONCLUSIONS}	
In this work, we propose a pseudo-label guided cross-video pixel contrast method for label-efficient surgical scene segmentation. 
The proposed method addresses the scarcity of annotated frames by providing a novel approach to utilize unlabeled video frames. 
We generate relatively accurate pseudo labels by sampling labeled frames equidistantly and then use pseudo labels as guidance to conduct pixel-level contrastive learning. To further improve the performance, we construct pairs in contrastive learning from video-wise. Our method achieves promising results on two real-world datasets EndoVis18 and CaDIS, and greatly reduces the burden of annotation. 
%Although we improve the overall segmentation performance, there are some classes with low performance due to class imbalance. 
In our future work, we shall explore how to solve class imbalance problem by introducing re-weighting and re-sampling strategies into contrastive learning. %~\cite{kang2019decoupling}
Moreover, we will investigate how to leverage the multi-modal resources, such as multi-view stereo images and depth information, to improve the performance. We will also investigate how to select frames as labelled ones based on active learning to further reduce labelling burden.

\bibliographystyle{IEEEtran}
\bibliography{mybibfile}

\begin{thebibliography}{10}
\providecommand{\url}[1]{#1}
\csname url@samestyle\endcsname
\providecommand{\newblock}{\relax}
\providecommand{\bibinfo}[2]{#2}
\providecommand{\BIBentrySTDinterwordspacing}{\spaceskip=0pt\relax}
\providecommand{\BIBentryALTinterwordstretchfactor}{4}
\providecommand{\BIBentryALTinterwordspacing}{\spaceskip=\fontdimen2\font plus
\BIBentryALTinterwordstretchfactor\fontdimen3\font minus
  \fontdimen4\font\relax}
\providecommand{\BIBforeignlanguage}[2]{{%
\expandafter\ifx\csname l@#1\endcsname\relax
\typeout{** WARNING: IEEEtran.bst: No hyphenation pattern has been}%
\typeout{** loaded for the language `#1'. Using the pattern for}%
\typeout{** the default language instead.}%
\else
\language=\csname l@#1\endcsname
\fi
#2}}
\providecommand{\BIBdecl}{\relax}
\BIBdecl

\bibitem{palep2009robotic}
J.~H. Palep, ``Robotic assisted minimally invasive surgery,'' \emph{Journal of
  minimal access surgery}, vol.~5, no.~1, p.~1, 2009.

\bibitem{maier2017surgical}
L.~Maier-Hein, S.~S. Vedula, S.~Speidel, N.~Navab, R.~Kikinis, A.~Park,
  M.~Eisenmann, H.~Feussner, G.~Forestier, S.~Giannarou \emph{et~al.},
  ``Surgical data science for next-generation interventions,'' \emph{Nat.
  Biomed. Eng.}, vol.~1, no.~9, pp. 691--696, 2017.

\bibitem{poursartip2018analysis}
B.~Poursartip, M.-E. LeBel, R.~V. Patel, M.~D. Naish, and A.~L. Trejos,
  ``Analysis of energy-based metrics for laparoscopic skills assessment,''
  \emph{IEEE Transactions on Biomedical Engineering}, vol.~65, no.~7, pp.
  1532--1542, 2018.

\bibitem{allan20202018}
M.~Allan, S.~Kondo, S.~Bodenstedt, S.~Leger, R.~Kadkhodamohammadi, I.~Luengo,
  F.~Fuentes, E.~Flouty, A.~Mohammed, M.~Pedersen \emph{et~al.}, ``2018 robotic
  scene segmentation challenge,'' \emph{arXiv preprint arXiv:2001.11190}, 2020.

\bibitem{allan20183}
M.~Allan, S.~Ourselin, D.~J. Hawkes, J.~D. Kelly, and D.~Stoyanov, ``3-d pose
  estimation of articulated instruments in robotic minimally invasive
  surgery,'' \emph{IEEE transactions on medical imaging}, vol.~37, no.~5, pp.
  1204--1213, 2018.

\bibitem{du2019patch}
X.~Du, M.~Allan, S.~Bodenstedt, L.~Maier-Hein, S.~Speidel, A.~Dore, and
  D.~Stoyanov, ``Patch-based adaptive weighting with segmentation and scale
  (pawss) for visual tracking in surgical video,'' \emph{Medical image
  analysis}, vol.~57, pp. 120--135, 2019.

\bibitem{nagy2019dvrk}
T.~D. Nagy and T.~Haidegger, ``A dvrk-based framework for surgical subtask
  automation,'' \emph{Acta Polytechnica Hungarica}, pp. 61--78, 2019.

\bibitem{ren2020task}
X.~Ren, S.~Ahmad, L.~Zhang, L.~Xiang, D.~Nie, F.~Yang, Q.~Wang, and D.~Shen,
  ``Task decomposition and synchronization for semantic biomedical image
  segmentation,'' \emph{IEEE Transactions on Image Processing}, vol.~29, pp.
  7497--7510, 2020.

\bibitem{grammatikopoulou2021cadis}
M.~Grammatikopoulou, E.~Flouty, A.~Kadkhodamohammadi, G.~Quellec, A.~Chow,
  J.~Nehme, I.~Luengo, and D.~Stoyanov, ``Cadis: Cataract dataset for surgical
  rgb-image segmentation,'' \emph{Medical Image Analysis}, vol.~71, p. 102053,
  2021.

\bibitem{wang2021noisy}
B.~Wang, L.~Li, Y.~Nakashima, R.~Kawasaki, H.~Nagahara, and Y.~Yagi,
  ``Noisy-lstm: Improving temporal awareness for video semantic segmentation,''
  \emph{IEEE Access}, vol.~9, pp. 46\,810--46\,820, 2021.

\bibitem{jin2022exploring}
Y.~Jin, Y.~Yu, C.~Chen, Z.~Zhao, P.-A. Heng, and D.~Stoyanov, ``Exploring
  intra-and inter-video relation for surgical semantic scene segmentation,''
  \emph{IEEE Transactions on Medical Imaging}, 2022.

\bibitem{lee2013pseudo}
D.-H. Lee \emph{et~al.}, ``Pseudo-label: The simple and efficient
  semi-supervised learning method for deep neural networks,'' in \emph{Workshop
  on challenges in representation learning, ICML}, vol.~3, no.~2, 2013, p. 896.

\bibitem{iscen2019label}
A.~Iscen, G.~Tolias, Y.~Avrithis, and O.~Chum, ``Label propagation for deep
  semi-supervised learning,'' in \emph{Proceedings of the IEEE/CVF Conference
  on Computer Vision and Pattern Recognition}, 2019, pp. 5070--5079.

\bibitem{jin2019incorporating}
Y.~Jin, K.~Cheng, Q.~Dou, and P.-A. Heng, ``Incorporating temporal prior from
  motion flow for instrument segmentation in minimally invasive surgery
  video,'' in \emph{International Conference on Medical Image Computing and
  Computer-Assisted Intervention}.\hskip 1em plus 0.5em minus 0.4em\relax
  Springer, 2019, pp. 440--448.

\bibitem{zhao2020learning}
Z.~Zhao, Y.~Jin, X.~Gao, Q.~Dou, and P.-A. Heng, ``Learning motion flows for
  semi-supervised instrument segmentation from robotic surgical video,'' in
  \emph{International Conference on Medical Image Computing and
  Computer-Assisted Intervention}.\hskip 1em plus 0.5em minus 0.4em\relax
  Springer, 2020, pp. 679--689.

\bibitem{zou2018unsupervised}
Y.~Zou, Z.~Yu, B.~Kumar, and J.~Wang, ``Unsupervised domain adaptation for
  semantic segmentation via class-balanced self-training,'' in
  \emph{Proceedings of the European conference on computer vision (ECCV)},
  2018, pp. 289--305.

\bibitem{hung2018adversarial}
W.-C. Hung, Y.-H. Tsai, Y.-T. Liou, Y.-Y. Lin, and M.-H. Yang, ``Adversarial
  learning for semi-supervised semantic segmentation,'' \emph{arXiv preprint
  arXiv:1802.07934}, 2018.

\bibitem{rizve2021defense}
M.~N. Rizve, K.~Duarte, Y.~S. Rawat, and M.~Shah, ``In defense of
  pseudo-labeling: An uncertainty-aware pseudo-label selection framework for
  semi-supervised learning,'' \emph{arXiv preprint arXiv:2101.06329}, 2021.

\bibitem{feng2020dmt}
Z.~Feng, Q.~Zhou, Q.~Gu, X.~Tan, G.~Cheng, X.~Lu, J.~Shi, and L.~Ma, ``Dmt:
  Dynamic mutual training for semi-supervised learning,'' \emph{arXiv preprint
  arXiv:2004.08514}, 2020.

\bibitem{french2019semi}
G.~French, S.~Laine, T.~Aila, M.~Mackiewicz, and G.~Finlayson,
  ``Semi-supervised semantic segmentation needs strong, varied perturbations,''
  \emph{arXiv preprint arXiv:1906.01916}, 2019.

\bibitem{tarvainen2017mean}
A.~Tarvainen and H.~Valpola, ``Mean teachers are better role models:
  Weight-averaged consistency targets improve semi-supervised deep learning
  results,'' \emph{arXiv preprint arXiv:1703.01780}, 2017.

\bibitem{yu2019uncertainty}
L.~Yu, S.~Wang, X.~Li, C.-W. Fu, and P.-A. Heng, ``Uncertainty-aware
  self-ensembling model for semi-supervised 3d left atrium segmentation,'' in
  \emph{International Conference on Medical Image Computing and
  Computer-Assisted Intervention}.\hskip 1em plus 0.5em minus 0.4em\relax
  Springer, 2019, pp. 605--613.

\bibitem{chen2020simple}
T.~Chen, S.~Kornblith, M.~Norouzi, and G.~Hinton, ``A simple framework for
  contrastive learning of visual representations,'' in \emph{International
  conference on machine learning}.\hskip 1em plus 0.5em minus 0.4em\relax PMLR,
  2020, pp. 1597--1607.

\bibitem{chen2020improved}
X.~Chen, H.~Fan, R.~Girshick, and K.~He, ``Improved baselines with momentum
  contrastive learning,'' \emph{arXiv preprint arXiv:2003.04297}, 2020.

\bibitem{chaitanya2020contrastive}
K.~Chaitanya, E.~Erdil, N.~Karani, and E.~Konukoglu, ``Contrastive learning of
  global and local features for medical image segmentation with limited
  annotations,'' \emph{arXiv preprint arXiv:2006.10511}, 2020.

\bibitem{lai2021semi}
X.~Lai, Z.~Tian, L.~Jiang, S.~Liu, H.~Zhao, L.~Wang, and J.~Jia,
  ``Semi-supervised semantic segmentation with directional context-aware
  consistency,'' in \emph{Proceedings of the IEEE/CVF Conference on Computer
  Vision and Pattern Recognition}, 2021, pp. 1205--1214.

\bibitem{xie2021propagate}
Z.~Xie, Y.~Lin, Z.~Zhang, Y.~Cao, S.~Lin, and H.~Hu, ``Propagate yourself:
  Exploring pixel-level consistency for unsupervised visual representation
  learning,'' in \emph{Proceedings of the IEEE/CVF Conference on Computer
  Vision and Pattern Recognition}, 2021, pp. 16\,684--16\,693.

\bibitem{zhao2020contrastive}
X.~Zhao, R.~Vemulapalli, P.~Mansfield, B.~Gong, B.~Green, L.~Shapira, and
  Y.~Wu, ``Contrastive learning for label-efficient semantic segmentation,''
  \emph{arXiv preprint arXiv:2012.06985}, 2020.

\bibitem{zhou2021c3}
Y.~Zhou, H.~Xu, W.~Zhang, B.~Gao, and P.-A. Heng, ``C3-semiseg: Contrastive
  semi-supervised segmentation via cross-set learning and dynamic
  class-balancing,'' in \emph{Proceedings of the IEEE/CVF International
  Conference on Computer Vision}, 2021, pp. 7036--7045.

\bibitem{wang2021exploring}
W.~Wang, T.~Zhou, F.~Yu, J.~Dai, E.~Konukoglu, and L.~Van~Gool, ``Exploring
  cross-image pixel contrast for semantic segmentation,'' in \emph{Proceedings
  of the IEEE/CVF International Conference on Computer Vision}, 2021, pp.
  7303--7313.

\bibitem{he2020momentum}
K.~He, H.~Fan, Y.~Wu, S.~Xie, and R.~Girshick, ``Momentum contrast for
  unsupervised visual representation learning,'' in \emph{Proceedings of the
  IEEE/CVF Conference on Computer Vision and Pattern Recognition}, 2020, pp.
  9729--9738.

\bibitem{wei2020theoretical}
C.~Wei, K.~Shen, Y.~Chen, and T.~Ma, ``Theoretical analysis of self-training
  with deep networks on unlabeled data,'' \emph{arXiv preprint
  arXiv:2010.03622}, 2020.

\bibitem{hunter1986exponentially}
J.~S. Hunter, ``The exponentially weighted moving average,'' \emph{Journal of
  quality technology}, vol.~18, no.~4, pp. 203--210, 1986.

\bibitem{chen2017rethinking}
L.-C. Chen, G.~Papandreou, F.~Schroff, and H.~Adam, ``Rethinking atrous
  convolution for semantic image segmentation,'' \emph{arXiv preprint
  arXiv:1706.05587}, 2017.

\bibitem{he2016deep}
K.~He, X.~Zhang, S.~Ren, and J.~Sun, ``Deep residual learning for image
  recognition,'' in \emph{Proceedings of the IEEE conference on computer vision
  and pattern recognition}, 2016, pp. 770--778.

\bibitem{deng2009imagenet}
J.~Deng, W.~Dong, R.~Socher, L.-J. Li, K.~Li, and L.~Fei-Fei, ``Imagenet: A
  large-scale hierarchical image database,'' in \emph{2009 IEEE conference on
  computer vision and pattern recognition}.\hskip 1em plus 0.5em minus
  0.4em\relax Ieee, 2009, pp. 248--255.

\bibitem{shrivastava2016training}
A.~Shrivastava, A.~Gupta, and R.~Girshick, ``Training region-based object
  detectors with online hard example mining,'' in \emph{Proceedings of the IEEE
  conference on computer vision and pattern recognition}, 2016, pp. 761--769.

\bibitem{ruder2016overview}
S.~Ruder, ``An overview of gradient descent optimization algorithms,''
  \emph{arXiv preprint arXiv:1609.04747}, 2016.

\bibitem{you2017large}
Y.~You, I.~Gitman, and B.~Ginsburg, ``Large batch training of convolutional
  networks,'' \emph{arXiv preprint arXiv:1708.03888}, 2017.

\bibitem{van2008visualizing}
L.~Van~der Maaten and G.~Hinton, ``Visualizing data using t-sne.''
  \emph{Journal of machine learning research}, vol.~9, no.~11, 2008.

\end{thebibliography}

\end{document}